# FULLY-AUTOMATIC SEGMENTATION OF KIDNEYS IN CLINICAL ULTRASOUND IMAGES USING A BOUNDARY DISTANCE REGRESSION NETWORK

*Shi Yin[1,2], Zhengqiang Zhang[1], Hongming Li[2], Qinmu Peng[1], Xinge You[1], Susan L. Furth[3], Gregory E. Tasian[4,5,6], Yong Fan[2]*

[1]School of Electronic Information and Communications, Huazhong University of Science and Technology, Wuhan, China
[2]Department of Radiology, [6]Department of Biostatistics, Epidemiology, and Informatics, Perelman School of Medicine, University of Pennsylvania, Philadelphia, PA, 19104, USA
[3]Department of Pediatrics, Division of Pediatric Nephrology, [4]Department of Surgery, Division of Pediatric Urology, [5]Center for Pediatric Clinical Effectiveness, Children's Hospital of Philadelphia, Philadelphia, PA 19104, USA

## ABSTRACT

It remains challenging to automatically segment kidneys in clinical ultrasound images due to the kidneys' varied shapes and image intensity distributions, although semi-automatic methods have achieved promising performance. In this study, we developed a novel boundary distance regression deep neural network to segment the kidneys, informed by the fact that the kidney boundaries are relatively consistent across images in terms of their appearance. Particularly, we first use deep neural networks pre-trained for classification of natural images to extract high-level image features from ultrasound images, then these feature maps are used as input to learn kidney boundary distance maps using a boundary distance regression network, and finally the predicted boundary distance maps are classified as kidney pixels or non-kidney pixels using a pixel classification network in an end-to-end learning fashion. Experimental results have demonstrated that our method could effectively improve the performance of automatic kidney segmentation, significantly better than deep learning based pixel classification networks.

## 1. INTRODUCTION

Ultrasound (US) imaging has been widely used to aid diagnosis and prognosis of acute and chronic kidney diseases. In particular, anatomic characteristics derived from US imaging, such as renal elasticity, maximum renal length, and cortical thickness, are associated with kidney function and lower renal parenchymal area as measured on US imaging data is associated with increased risk of end-stage renal disease (ESRD) in boys with posterior urethral valves [1]. Imaging features computed from kidney US imaging data using deep convolutional neural networks (CNNs) have demonstrated improved classification performance for distinguishing children with congenital abnormalities of the kidney and urinary tract (CAKUT) from controls [2]. The computation of these anatomic measures typically involves manual or semi-automatic segmentation of the kidney in US images, which increases inter-operator variability and reduces reliability. Therefore, automatic and reliable segmentation of the kidney from US imaging data is desired.

Since manual segmentation of the kidney is time consuming, labor-intensive, and highly prone to intra- and inter-operator variability, semi-automatic and interactive segmentation methods have been developed. Particularly, an interactive tool has been developed for detecting and segmenting the kidney in 3D US images [3]. A semi-automatic segmentation framework based on both texture and shape priors has been proposed for segmenting the kidney from US images [4]. A novel graph cuts method has been proposed to segment the kidney in US images by integrating image intensity information and texture feature maps [5]. A variety of methods have been proposed to segment the kidney based on active shape models and statistical shape models [3, 6-9]. Random forests have also been adopted in a semi-automatic segmentation method to segment the kidney [10]. Despite the fact that different strategies have been adopted in the semi-automatic kidney segmentation methods, most of them solve the kidney segmentation problem as a boundary detection problem. Although these methods have achieved excellent kidney segmentation performance, most of them rely on manual operations for initializing the semi-automatic segmentation.

Deep CNNs have demonstrated excellent performance in a variety of image segmentation tasks, including semantic segmentation of natural images [11, 12] and medical image segmentation [13-16]. Recently, several methods have been proposed to automatically segment the kidney from medical imaging data to generate kidney masks based on deep CNNs. In particular, 2D and 3D U-net neural networks have been adopted to segment the kidney by classifying image pixels/voxels as kidney or non-kidney ones in a pattern classification setting [17-20]. In these pattern classification based kidney segmentation methods, all pixels/voxels within the kidney have the same kidney classification labels, ignoring large variability of the kidneys in both appearance and shape in US images (illustrated in Fig. 4). Such shape and appearance variability, in conjunction with inherent speckle noise of US images, may degrade performance of the pattern classification based kidney segmentation methods.

Inspired by the excellent performance of the semi-automatic boundary detection based kidney segmentation methods, we develop a fully automatic, end-to-end deep learning method to consecutively learn kidney boundaries and pixelwise kidney masks from a set of manually labeled US images. Instead of distinguishing kidney pixels from non-kidney ones in a pattern classification setting, we learn CNNs in a regression setting to detect kidney boundaries that are modeled as boundary distance maps. From the learned boundary distance maps, we subsequently learn pixelwise kidney masks by optimizing their overlap with the manual kidney segmentation labels. To augment the training dataset, we adopt a kidney shape based image registration method to generate more training samples. Our deep CNNs are built upon an image segmentation network architecture derived from DeepLab [12] so that existing image classification/segmentation models could be reused as a starting point of the kidney image segmentation in a transfer learning framework to speed up the model training and improve the performance of the kidney image segmentation.

We have evaluated the proposed method for segmenting the kidney based on clinical US images, and the evaluation results have demonstrated that the proposed method could achieve promising segmentation performance and outperformed alternative state-of-the-art deep learning based image segmentation methods.

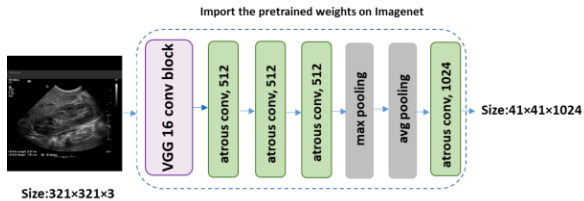

**Fig. 1.** Network architecture of the feature extraction network. We import weights of VGG-16 pre-trained on the Imagenet.

## 2. METHODS

### 2.1 Transfer learning for extracting image features

Instead of directly building a boundary detection network on raw US images, we adopt transfer learning techniques to extract informative high level image features from US images as a starting point. Particularly, we extract image features from US images by utilizing a general image classification network VGG-16. As illustrated in Fig. 1, we employed the VGG-16 network pre-trained on the Imagenet to learn image features for the kidney segmentation. To fine-tune the image features for the kidney segmentation, we follow the Deeplab architecture by applying atrous convolutions to compute dense image feature representations. As illustrated in Fig. 1, after atrous convolution (conv) and pooling layers, a feature map of $41 \times 41 \times 1024$ is obtained from each input image of $321 \times 321$. The input image is duplicated 3 times to generate a pseudo-color image so that the pre-trained VGG-16 network could be used to learn image features from US images in a transfer learning setting.

.

### 2.2. Boundary distance regression network

Since kidneys have large variability in both appearance and shape in clinical US images as illustrated in Fig. 4, we develop a boundary distance regression network, as illustrated in Fig. 2, for detecting the boundary of kidney, instead of building a pixelwise classification network for classifying pixels into kidney or non-kidney pixels (suffering from large variability) or classifying pixels into kidney boundary or non-boundary pixels (suffering from unbalanced samples). The boundary distance regression network mainly consists of two parts: a projection part that produces boundary feature maps and a high-resolution reconstruction part that upsamples the feature maps to obtain the kidney distance maps at the same spatial resolution of the input raw image. The projection part is built on convolutional layers, and the reconstruction part is built on deconvolution layers. The deconvolution operation of the $i$th layer is defined as
$$S_i = \max(0, W_i \otimes S_{i-1} + B_i),$$
where $W_i$ contains $n_i$ filters of size $f_i \times f_i$, $B_i$ is a $n_i$-dimensional vector, and $\otimes$ is deconvolution operator. The upsampling deconvolution layers double the spatial dimension of their input feature maps, and therefore 3 upsampling deconvolution layers are adopted in the kidney boundary regression network to learn the kidney boundary in the input image space.

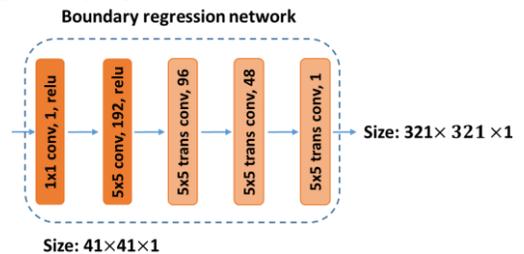

**Fig. 2.** Network architecture of the boundary regression network.

Given a training US image $I$ with its kidney boundary, we compute the distance to the kidney boundary for every pixel $P_i \in I$ of the input image and obtain a normalized kidney distance map of the same size of the input image using potential function as following:
$$d_{P_i} = e^{-D_i}$$
where $D_i = \min_{b_j \in \boldsymbol{b}} \text{dist}(Pi, b_j)$ is the minimal Euclidean distance of pixel $P_i$ to the kidney boundary pixels $\boldsymbol{b} = \{b_j\}_{j \in J}$. As illustrated in Fig.3, pixels on the kidney boundary have normalized exponential kidney distance equal to 1. The training distance maps are used to optimize the boundary distance regression network, which is then applied to testing input images to obtain their kidney distance maps. Euclidean distance between the predicted and real distance maps was used as the loss function.

To obtain a smooth closed contour of the kidney boundary, we construct a minimum spanning tree of all predicted kidney boundary pixels. We first use a threshold to binarize the obtained distance map, then construct an undirected graph to identify its minimum spanning trees, and finally the max path of the minimum spanning tree is obtained as a close contour of the kidney boundary and a binary mask of the kidney is subsequently obtained. We refer to the boundary distance regression network followed by the aforementioned post-processing for segmenting kidneys as a boundary regression network (BRN) hereafter.

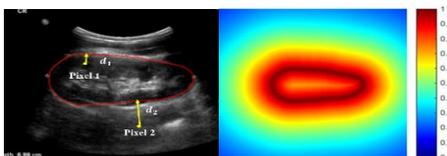

**Fig. 3.** A kidney image with manually labeled boundary (left) and an example kidney distance map (right).

### 2.3. Kidney mask segmentation network

To obtain kidney masks from the predicted kidney distance maps in an end-to-end fashion, we further train a kidney mask segmentation network that is essentially a pixelwise segmentation network to classify pixels into kidney or non-kidney ones based on their kidney distance measures. Specifically, the same network architecture used in the kidney boundary regression network is adopted for the classification network, with following modifications: 1) the input is the predicted distance map and 2) the cost function is a softmax loss based classification cost function.

### 2.4. End-to-end training of deep learning networks

The softmax loss function and the regression network loss function are combined to train the boundary regression network and mask segmentation network in an end-to-end fashion. More weight is put on the kidney boundary regression cost function in the early stage of the network training, and then shifts to the kidney pixel classification cost function in the late stage. The training data were augmented by deformable image registration between pairs of training images. The kidney pixel classification network's output is adopted as the final segmentation result.

## 3. EXPERIMENTAL RESULTS

We validated our method based on kidney images collected at the Children's Hospital of Philadelphia (CHOP). The dataset contains 185 kidney images obtained from 50 normal subjects and 50 patients. The abnormal kidney images are from the children with CAKUT. Particularly, all the 50 normal subjects had both left and right kidney images, 35 patients had abnormal kidneys in both sides, 4 patients had abnormal kidneys on the left side only, and 11 patients had abnormal kidneys on the right side only. We randomly select 105 kidney images as training data and 80 kidney images as testing data. The boundaries of the kidneys were annotated by experts from the CHOP.

We compare our method with the FCN and Deeplab [11, 12], which were fine-tuned using the same training data without data augmentation. The two methods were designed to segment images by classifying pixels. The segmentation performance was evaluated using dice coefficient, mean distance index, and accuracy index [5]. We also reported results obtained by the proposed boundary regression network only.

Table 1: Segmentation performance of methods under comparison.

|  | Dice coefficient | | Mean distance | | Accuracy | |
| --- | --- | --- | --- | --- | --- | --- |
|  | Mean±Std | p-value | Mean±Std | p-value | Mean±Std | p-value |
| FCN | 0.78±0.11 | 9e-12 | 9.83±4.82 | 4e-14 | 0.96±0.02 | 1e-11 |
| Deeplab | 0.86±0.11 | 4e-9 | 5.85±3.28 | 2e-11 | 0.97±0.02 | 3e-8 |
| BRN | 0.93±0.07 | - | 3.42±2.28 | - | 0.987±0.009 | - |
| Proposed | 0.94±0.03 | - | 2.72±1.61 | - | 0.989±0.006 | - |

As summarized in Table 1, the proposed method outperformed the alternative methods under comparison with statistical significance (Wilcoxon signed-rank test), and the proposed end-to-end learning method had better performance than the boundary regression network. Fig. 4 shows example segmentation results, indicating that our method was robust to kidney appearance variability. However, the alternative methods under comparison had worse performance for kidney images with large appearance variability, partially due to the fact that the direct classification of pixels was not robust to the appearance variability. On average, it took 0.18s to segment one US image on a GeForce 1060 GPU.

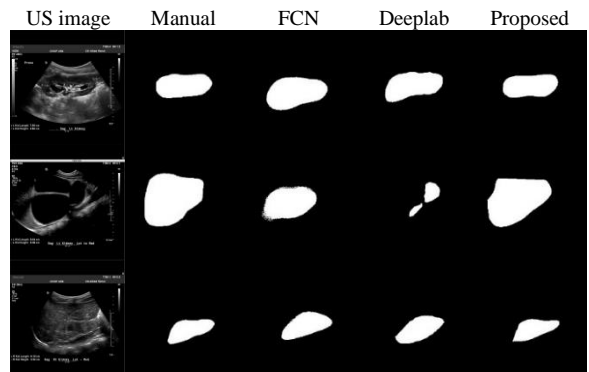

**Fig. 4.** Example US images with heterogenous appearance, manual kidney segmentation results, and kidney segmentation results obtained by the methods under comparison.

Fig. 5 shows representative segmentation results obtained by the kidney boundary regression network and the end-to-end subsequent boundary distance regression and pixelwise classification networks, demonstrating that the end-to-end learning could obtain better performance for kidneys with blurring boundaries. More importantly, we can obtain the kidney masks from their distance maps without any post-processing step.

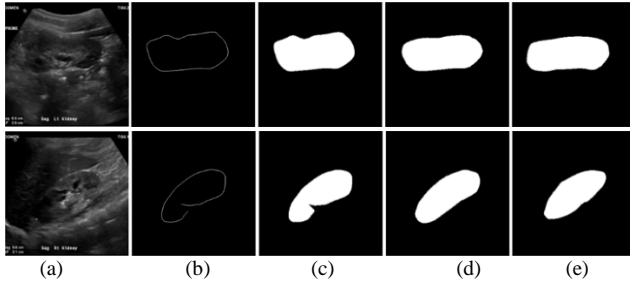

(a) (b) (c) (d) (e)

**Fig. 5.** Results for the boundary regression network and the end-to-end learning networks. (a) input kidney US images, (b) binary skeleton maps of the predicted distance maps, (c) kidney masks obtained with the minimum spanning tree based post-processing, (d) kidney masks obtained by the end-to-end subsequent segmentation network, and (e) kidney masks obtained by manual labels.

## 4. CONCLUSIONS

In this study, we proposed a novel boundary regression network architecture to achieve fully-automatic kidney segmentation. Our method obtained promising segmentation performance for segmenting clinical kidney US images with large variability in both appearance and shape, which is significantly better than the alternatives under comparison, demonstrating the boundary detection strategy works better than standard classification techniques for clinical US images with heterogenous appearance. We also demonstrated that the end-to-end learning strategy could further improve the boundary distance regression network.

## ACKNOWLEDGEMENTS

This work was supported in part by the National Institutes of Health grant (DK114786); the National Center for Advancing Translational Sciences of the National Institutes of Health (UL1TR001878); the National Natural Science Foundation of China grant (61772220); the Key Program for International S&T Cooperation Projects of China (2016YFE0121200); the Hubei Province Technological Innovation Major Project (2017AAA017 and 2018ACA135); the Institute for Translational Medicine and Therapeutics' (ITMAT) Transdisciplinary Program in Translational Medicine and Therapeutics, and the China Scholarship Council.

## REFERENCES

[1] J. E. Pulido *et al.*, "Renal parenchymal area and risk of ESRD in boys with posterior urethral valves," *Clin J Am Soc Nephrol,* vol. 9, no. 3, pp. 499-505, Mar 2014.

[2] Q. Zheng *et al.*, "Transfer learning for diagnosis of congenital abnormalities of the kidney and urinary tract in children based on Ultrasound imaging data," *IEEE International Symposium on Biomedical Imaging (ISBI),* pp. 1487-1490, April 2018.

[3] R. Ardon *et al.*, "Fast kidney detection and segmentation with learned kernel convolution and model deformation in 3D ultrasound images," in *2015 IEEE 12th International Symposium on Biomedical Imaging (ISBI)*, 2015, pp. 268-271.

[4] X. Jun *et al.*, "Segmentation of kidney from ultrasound images based on texture and shape priors," *IEEE Transactions on Medical Imaging,* vol. 24, no. 1, pp. 45-57, 2005.

[5] Q. Zheng *et al.*, "A Dynamic Graph Cuts Method with Integrated Multiple Feature Maps for Segmenting Kidneys in 2D Ultrasound Images," *Acad Radiol,* Feb 12 2018.

[6] M. Martín-Fernández and C. Alberola-López, "An approach for contour detection of human kidneys from ultrasound images using Markov random fields and active contours," *Medical Image Analysis,* vol. 9, no. 1, pp. 1-23, 2005/02/01/ 2005.

[7] J. J. Cerrolaza *et al.*, "Segmentation of kidney in 3D-ultrasound images using Gabor-based appearance models," in *2014 IEEE 11th International Symposium on Biomedical Imaging (ISBI)*, 2014, pp. 633-636.

[8] C. S. Mendoza *et al.*, "Automatic Analysis of Pediatric Renal Ultrasound Using Shape, Anatomical and Image Acquisition Priors," in *Medical Image Computing and Computer-Assisted Intervention – MICCAI 2013*, pp. 259-266, 2013

[9] J. J. Cerrolaza *et al.*, "Renal Segmentation From 3D Ultrasound via Fuzzy Appearance Models and Patient-Specific Alpha Shapes," *IEEE Transactions on Medical Imaging,* vol. 35, no. 11, pp. 2393-2402, 2016.

[10] K. Sharma *et al.*, "Semi-automatic segmentation of autosomal dominant polycystic kidneys using random forests," *arXiv:1510.06915,* 2015.

[11] J. Long *et al.*, "Fully convolutional networks for semantic segmentation," presented at the The IEEE Conference on Computer Vision and Pattern Recognition, 2015.

[12] L.-C. Chen *et al.*, "Deeplab: Semantic image segmentation with deep convolutional nets, atrous convolution, and fully connected crfs," *IEEE Transactions on Pattern Analysis and Machine Intelligence,* vol. 40, no. 4, pp. 834-848, 2018.

[13] O. Ronneberger *et al.*, "U-Net: Convolutional Networks for Biomedical Image Segmentation," in *Medical Image Computing and Computer-Assisted Intervention – MICCAI 2015*, pp. 234-241, 2015.

[14] X. Zhao *et al.*, "A deep learning model integrating FCNNs and CRFs for brain tumor segmentation," *Medical Image Analysis,* vol. 43, pp. 98-111, 2018.

[15] C. Li *et al.*, "DeepMitosis: Mitosis detection via deep detection, verification and segmentation networks," *Medical Image Analysis,* vol. 45, pp. 121-133, 2018/04/01/ 2018.

[16] X. Zhao *et al.*, "3D Brain Tumor Segmentation Through Integrating Multiple 2D FCNNs," 2018, pp. 191-203: Springer International Publishing.

[17] K. Sharma *et al.*, "Automatic segmentation of kidneys using deep learning for total kidney volume quantification in autosomal dominant polycystic kidney disease," *Scientific reports,* vol. 7, no. 1, p. 2049, 2017.

[18] P. Jackson *et al.*, "Deep Learning Renal Segmentation for Fully Automated Radiation Dose Estimation in Unsealed Source Therapy," *Frontiers in Oncology,* vol. 8, p. 215, 2018.

[19] H. Ravishankar *et al.*, "Joint Deep Learning of Foreground, Background and Shape for Robust Contextual Segmentation," in *Information Processing in Medical Imaging: 25th International Conference, IPMI 2017,* pp. 622-632.

[20] H. Ravishankar *et al.*, "Learning and Incorporating Shape Models for Semantic Segmentation," in *Medical Image Computing and Computer Assisted Intervention − MICCAI 2017*, pp. 203-211.